# INDUCING FEATURES OF RANDOM FIELDS[*]

Stephen Della Pietra[†]  Vincent Della Pietra[†]  John Lafferty[‡]

May, 1995

CMU-CS-95-144

[†]IBM Thomas J. Watson Research Center
P.O. Box 704
Yorktown Heights, NY 10598

[‡]School of Computer Science
Carnegie Mellon University
Pittsburgh, PA 15213

[*] Correspondence regarding this paper should be sent to John Lafferty at the above address, or by e-mail to lafferty@cs.cmu.edu.
[†] Research supported in part by ARPA under grant N00014-91-C-0135.
[‡] Research supported in part by NSF and ARPA under grants IRI-9314969 and N00014-92-C-0189.

The views and conclusions contained in this document are those of the authors and should not be interpreted as representing the official policies, either expressed or implied, of the NSF or the U.S. government.

**Abstract**

We present a technique for constructing random fields from a set of training samples. The learning paradigm builds increasingly complex fields by allowing potential functions, or features, that are supported by increasingly large subgraphs. Each feature has a weight that is trained by minimizing the Kullback-Leibler divergence between the model and the empirical distribution of the training data. A greedy algorithm determines how features are incrementally added to the field and an iterative scaling algorithm is used to estimate the optimal values of the weights.

The random field models and techniques introduced in this paper differ from those common to much of the computer vision literature in that the underlying random fields are non-Markovian and have a large number of parameters that must be estimated. Relations to other learning approaches including decision trees and Boltzmann machines are given. As a demonstration of the method, we describe its application to the problem of automatic word classification in natural language processing.

# 1. Introduction

In this paper we present a method for incrementally constructing random fields. Our method builds increasingly complex fields to approximate the empirical distribution of a set of training examples by allowing potential functions, or features, that are supported by increasingly large subgraphs. Each feature is assigned a weight, and the weights are trained to minimize the Kullback-Leibler divergence between the field and the empirical distribution of the training data. Features are incrementally added to a field using a top-down greedy algorithm.

To illustrate the nature of our approach, suppose that we have a set of images we wish to characterize by a statistical model. Each image is represented by an assignment of one of the colors red, blue, or green to the vertices of a square, 2-dimensional grid. How should the statistical model be constructed?

To begin, suppose we observe that vertices of the images are 50% red, 30% blue and 20% green. This leads us to characterize the statistical model in terms of the average number of vertices of each color, in a distribution of the form

$$p(\omega) = \frac{1}{Z} e^{\sum_i \lambda_r \delta(\omega_i, red) + \lambda_b \delta(\omega_i, blue) + \lambda_g \delta(\omega_i, green)} \tag{1.1}$$

where $\omega_i$ is the color of vertex $i$ in the image $\omega$. The weights $\lambda_r, \lambda_b, \lambda_g$ are chosen to reflect our observations of the frequencies of colors of individual vertices in the set of images. The more detailed observation that a red vertex is only rarely adjacent to a green vertex might then be included as a refinement to the model, leading to a distribution of the form

$$p'(\omega) = \frac{1}{Z'} e^{\sum_{i \sim j} \lambda_{r,g} \delta(\omega_i, red) \delta(\omega_j, green) + \sum_i \lambda_r \delta(\omega_i, red) + \lambda_b \delta(\omega_i, blue) + \lambda_g \delta(\omega_i, green)} \tag{1.2}$$

where the weight $\lambda_{r,g}$ is adjusted to reflect our specific observations on the colors of adjacent vertices, and any necessary readjustments are made to the weights $\lambda_r, \lambda_b, \lambda_g$ to respect our earlier observations. At the expense of an increasing number of parameters that need to be adjusted, an increasingly detailed set of features of the images can be characterized by the distribution. But which features should the model characterize, and how should the weights be chosen? In this paper we present a general framework for addressing these questions.

As another illustration, suppose we wish to automatically characterize spellings of words according to a statistical model; this is the application we develop in Section 5. A field with no features is simply a uniform distribution on ASCII strings (where we take the distribution of string *lengths* as given). The most conspicuous feature of English spellings is that they are most commonly comprised of lower-case letters. The induction algorithm makes this observation by first constructing the field

$$p(\omega) = \frac{1}{Z} e^{\sum_i \lambda_{[a-z]} \chi_{[a-z]}(\omega_i)}$$

where $\chi$ is an indicator function and the weight $\lambda_{[a-z]}$ associated with the feature that a character is lower-case is chosen to be approximately 1.944. This means that a string with a lowercase letter in some position is about $7 \approx e^{1.944}$ times more likely than the same string without a lowercase letter in that position. The following collection of strings was generated from the resulting field by Gibbs sampling:

```
m, r, xevo, ijjiir, b, to, jz, gsr, wq, vf, x, ga, msmGh,
pcp, d, oziVlal, hzagh, yzop, io, advzmxnv, ijv_bolft, x,
emx, kayerf, mlj, rawzyb, jp, ag, ctdnnnbg, wgdw, t, kguv,
cy, spxcq, uzflbbf, dxtkkn, cxwx, jpd, ztzh, lv, zhpkvnu,
l^, r, qee, nynrx, atze4n, ik, se, w, lrh, hp+, yrqyka'h,
zcngotcnx, igcump, zjcjs, lqpWiqu, cefmfhc, o, lb, fdcY, tzby,
yopxmvk, by, fz,, t, govyccm, ijyiduwfzo, 6xr, duh, ejv, pk,
pjw, l, fl, w
```

The second most important feature, according to the algorithm, is that two adjacent lower-case characters are extremely common. Accordingly, the second-order field becomes

$$p'(\omega) = \frac{1}{Z'} e^{\sum_{i \sim j} \lambda_{[a-z][a-z]} \chi_{[a-z]}(\omega_i) \chi_{[a-z]}(\omega_j) + \sum_i \lambda_{[a-z]} \chi_{[a-z]}(\omega_i)}$$

where the weight $\lambda_{[a-z][a-z]}$ associated with adjacent lower-case letters is approximately 1.80.

The first 1000 features that the algorithm induces include the strings s>, <re, ly>, and ing>, where the character "<" denotes beginning-of-string and the character ">" denotes end-of-string. In addition, the first 1000 features include the regular expressions [0-9][0-9] (with weight 9.15) and [a-z][A-Z] (with weight −5.81) in addition to the first two features [a-z] and [a-z][a-z]. A set of strings obtained by Gibbs sampling from the resulting field is shown below:

```
was, reaser, in, there, to, will, ,, was, by, homes, thing,
be, reloverated, ther, which, conists, at, fores, anditing, with,
Mr., proveral, the, ,, ***, on't, prolling, prothere, ,, mento,
at, yaou, 1, chestraing, for, have, to, intrally, of, qut, .,
best, compers, ***, cluseliment, uster, of, is, deveral, this,
thise, of, offect, inatever, thifer, constranded, stater, vill,
in, thase, in, youse, menttering, and, ., of, in, verate, of, to
```

These examples are discussed in detail in Section 5.

The induction algorithm that we present has two parts: *feature selection* and *parameter estimation*. The greediness of the algorithm arises in feature selection. In this step each feature in a pool of candidate features is evaluated by estimating the reduction in the Kullback-Leibler divergence that would result from adding the feature to the field. This reduction is approximated as a function of a single parameter, and the largest value of this function is called the *gain* of the candidate. The candidate with the largest gain is added to the field. In the parameter estimation step, the parameters of the field are estimated using an iterative scaling algorithm. The algorithm we use is a new statistical estimation algorithm that we call *Improved Iterative Scaling*. It is an improvement of the Generalized Iterative Scaling algorithm of Darroch and Ratcliff [19] in that it does not require that the features sum to a constant. The improved algorithm is easier to implement than the Darroch and Ratcliff algorithm, and can lead to an increase in the rate of convergence by increasing the size of the step taken toward the maximum at each iteration. In Section 4 we give a simple, self-contained proof of the convergence of the improved algorithm that does not make use of the Kuhn-Tucker theorem or other machinery of constrained optimization. Moreover, our proof does not rely on the convergence of alternating I-projection as in Csiszár's proof [16] of the Darroch-Ratcliff procedure.

Both the feature selection step and the parameter estimation step require the solution of certain algebraic equations whose coefficients are determined as expectation values with respect to the field. In many applications these expectations cannot be computed exactly because they involve a sum over an exponentially large number of configurations. This is true of the application that we develop in Section 5. In such cases it is possible to approximate the equations that must be solved using Monte Carlo techniques to compute expectations of random variables. The application that we present uses Gibbs sampling to compute expectations, and the resulting equations are then solved using Newton's method.

Our method can be viewed in terms of the *principle of maximum entropy* [26], which instructs us to assume an exponential form for our distributions, with the parameters viewed as Lagrange multipliers. The techniques that we develop in this paper apply to exponential models in general. We formulate our approach in terms of random fields because this provides a convenient framework within which to work, and because our main application is naturally cast in these terms.

Our method differs from the most common applications of statistical techniques in computer vision and natural language processing. In contrast to many applications in computer vision, which involve only a few free parameters, the typical application of our method involves the estimation of thousands of free parameters. In addition, our methods apply to general exponential models and random fields–there is no underlying Markov assumption made. In contrast to the statistical techniques common to natural language processing, in typical applications of our method there is no probabilistic finite-state or push-down automaton on which the statistical model is built.

In the following section we describe the form of the random field models considered in this paper and the general learning algorithm. In Section 3 we discuss the feature selection step of the algorithm and briefly address cases when the equations need to be estimated using Monte Carlo methods. In Section 4 we present the Improved Iterative Scaling algorithm for estimating the parameters, and prove the convergence of this algorithm. In Section 5 we present the application of inducing features of spellings, and finally in Section 6 we discuss the relation between our methods and other learning approaches, as well as possible extensions of our method.

## 2. The Learning Paradigm

In this section we present the basic algorithm for building up a random field from elementary features. The basic idea is to incrementally construct an increasingly detailed field to approximate a reference distribution $\tilde{p}$. Typically the distribution $\tilde{p}$ is obtained as the empirical distribution of a set of training examples. After establishing our notation and defining the form of the random field models we consider, we present the training problem as a statement of two equivalent optimization problems. We then discuss the notions of a candidate feature and the gain of a candidate. Finally, we give a statement of the induction algorithm.

*2.1 Form of the random field models.* Let $G = (E, V)$ be a finite graph with vertex set $V$

and edge set $E$, and let $\mathcal{A}$ be a finite alphabet. The *configuration space* $\Omega$ is the set of all labelings of the vertices in $V$ by letters in $\mathcal{A}$. If $C \subset V$ and $\omega \in \Omega$ is a configuration, then $\omega_C$ denotes the configuration restricted to $C$. A *random field* on $G$ is a probability distribution on $\Omega$. The set of all random fields is nothing more than the simplex $\Delta$ of all probability distributions on $\Omega$. If $f : \Omega \to \mathbf{R}$ then the *support* of $f$, written $\text{supp}(f)$, is the smallest vertex subset $C \subset V$ having the property that whenever $\omega, \omega' \in \Omega$ with $\omega_C = \omega'_C$ then $f(\omega) = f(\omega')$.

We consider random fields that are given by Gibbs distributions of the form

$$p(\omega) = \frac{1}{Z} e^{\sum_C V_C(\omega)} \tag{2.1}$$

for $\omega \in \Omega$, where $V_C : \Omega \to \mathbf{R}$ are functions with $\text{supp}(V_C) = C$. The field is *Markov* if whenever $V_C \neq 0$ then $C$ is a *clique*, or totally connected subset of $V$. This property is expressed in terms of conditional probabilities as

$$p(\omega_u \,|\, \omega_v, \ v \neq u) = p(\omega_u \,|\, \omega_v, \ (u,v) \in E) \tag{2.2}$$

where $u$ and $v$ are arbitrary vertices. We assume that each $C$ is a path-connected subset of $V$ and that

$$V_C(\omega) = \sum_{1 \leq i \leq n_C} \lambda_i^C f_i^C(\omega) = \lambda^C \cdot f^C(\omega) \tag{2.3}$$

where $\lambda_i^C \in \mathbf{R}$ and $f_i^C(\omega) \in \{0,1\}$. We say that the values $\lambda_i^C$ are the *parameters* of the field and that the functions $f_i^C$ are the *features* of the field. In the following, it will often be convenient to use notation that disregards the dependence of the features and parameters on a vertex subset $C$, expressing the field in the form

$$p(\omega) = \frac{1}{Z} e^{\sum_i \lambda_i f_i(\omega)} = \frac{1}{Z} e^{\lambda \cdot f(\omega)} . \tag{2.4}$$

For every random field $(E, V, \{\lambda_i, f_i\})$ of the above form, there is a field $(E', V, \{\lambda_i, f_i\})$ that is Markovian, obtained by completing the edge set $E$ to ensure that for each $i$, the subgraph generated by the vertex subset $C = \text{supp}(f_i)$ is totally connected.

If we impose the constraint $\lambda_i = \lambda_j$ on two parameters $\lambda_i$ and $\lambda_j$, then we say that these parameters are *tied*. If $\lambda_i$ and $\lambda_j$ are tied, then we can write

$$\lambda_i f_i(\omega) + \lambda_j f_j(\omega) = \lambda g(\omega) \tag{2.5}$$

where $g = f_i + f_j$ is a *non-binary* feature. In general, we can collapse any number of tied parameters onto a single parameter associated with a non-binary feature. Having tied parameters is often natural for a particular problem, but the presence of non-binary features generally makes the estimation of parameters more difficult.

A random field $(E, V, \{\lambda_i, f_i\})$ is said to have *homogeneous features* if for each feature $f_i$ and automorphism $\sigma$ of the graph $G = (E, V)$, there is a feature $f_j$ such that $f_j(\sigma\omega) = f_i(\omega)$ for $\omega \in \Omega$. If in addition $\lambda_j = \lambda_i$, then the field is said to be *homogeneous*. Homogeneous features arise naturally in the application of Section 5.

The methods that we describe in this paper apply to exponential models in general; that is, it is not essential that there is an underlying graph structure. However, it will be convenient to express our approach in terms of the random field models described above.

*2.2 Two optimization problems.* Suppose that we are given an initial model $q_0 \in \Delta$, a reference distribution $\tilde{p}$, and a set of features $f = (f_0, f_1, \ldots, f_n)$. In practice, it is often the case that $\tilde{p}$ is the empirical distribution of a set of training samples $\omega^{(1)}, \omega^{(2)} \ldots \omega^{(N)}$, and is thus given by

$$\tilde{p}(\omega) = \frac{c(\omega)}{N} \tag{2.6}$$

where $c(\omega) = \sum_{1 \leq i \leq N} \delta(\omega, \omega^{(i)})$ is the number of times that configuration $\omega$ appears among the training samples.

We wish to construct a probability distribution $q_\star \in \Delta$ that accounts for these data, in the sense that it approximates $\tilde{p}$ but does not deviate too far from $q_0$. We measure distance between probability distributions $p$ and $q$ in $\Delta$ using the Kullback-Leibler divergence

$$D(\tilde{p} \parallel p) = \sum_{\omega \in \Omega} \tilde{p}(\omega) \log \frac{\tilde{p}(\omega)}{p(\omega)} . \tag{2.7}$$

Throughout this paper we use the notation

$$p[g] = \sum_{\omega \in \Omega} g(\omega) p(\omega)$$

for the expectation of a function $g : \Omega \to \mathbf{R}$ with respect to the probability distribution $p$. For a function $h : \Omega \to \mathbf{R}$ and a distribution $q$, we use both the notation $h \circ q$ and $q_h$ to denote the generalized Gibbs distribution given by

$$q_h(\omega) = (h \circ q)(\omega) = \frac{1}{Z_q(h)} e^{h(\omega)} q(\omega) .$$

Note that $Z_q(h)$ is not the usual partition function. It is a normalization constant determined by the requirement that $(h \circ q)(\omega)$ sums to 1 over $\omega$, and can be written as an expectation:

$$Z_q(h) = q[e^h] .$$

There are two natural sets of probability distributions determined by the data $\tilde{p}$, $q_0$, and $f$. The first is the set $\mathcal{P}(f, \tilde{p})$ of all distributions that agree with $\tilde{p}$ as to the expected value of the feature function $f$:

$$\mathcal{P}(f, \tilde{p}) = \{p \in \Delta : p[f] = \tilde{p}[f]\}.$$

The second is the set $\mathcal{Q}(f, q_0)$ of generalized Gibbs distributions based on $q_0$ with feature function $f$:

$$\mathcal{Q}(f, q_0) = \{(\lambda \cdot f) \circ q_0 : \lambda \in \mathbf{R}^n\}.$$

We let $\bar{\mathcal{Q}}(f, q_0)$ denote the closure of $\mathcal{Q}(f, q_0)$ in $\Delta$ (with respect to the topology it inherits as a subset of Euclidean space).

There are two natural criteria for choosing $q_\star$:

- *Maximum Likelihood Gibbs Distribution.* Choose $q_\star$ to be a distribution in $\bar{\mathcal{Q}}(f, q_0)$ with maximum likelihood with respect to $\tilde{p}$:

$$q_\star = \underset{q \in \bar{\mathcal{Q}}(f, q_0)}{\arg\min}\, D(\tilde{p} \,\|\, q)$$

- *Maximum Entropy Constrained Distribution.* Choose $q_\star$ to be a distribution in $\mathcal{P}(f, \tilde{p})$ that has maximum entropy relative to $q_0$:

$$q_\star = \underset{p \in \mathcal{P}(f, \tilde{p})}{\arg\min}\, D(p \,\|\, q_0)$$

Although these criteria are different, they determine the same distribution. In fact, the following is true, as we prove in Section 4.

**Proposition.** *Suppose that $D(\tilde{p} \,\|\, q_0) < \infty$. Then there exists a unique $q_\star \in \Delta$ satisfying*

(1) $q_\star \in \mathcal{P}(f, \tilde{p}) \cap \bar{\mathcal{Q}}(f, q_0)$

(2) $D(p \,\|\, q) = D(p \,\|\, q_\star) + D(q_\star \,\|\, q)$ *for any* $p \in \mathcal{P}(f, \tilde{p})$ *and* $q \in \bar{\mathcal{Q}}(f, q_0)$

(3) $q_\star = \underset{q \in \bar{\mathcal{Q}}(f, q_0)}{\arg\min}\, D(\tilde{p} \,\|\, q)$

(4) $q_\star = \underset{p \in \mathcal{P}(f, \tilde{p})}{\arg\min}\, D(p \,\|\, q_0)$.

*Moreover, any of these four properties determines $q_\star$ uniquely.*

When $\tilde{p}$ is the empirical distribution of a set of training examples $\omega^{(1)}, \omega^{(2)} \ldots \omega^{(N)}$, minimizing $D(\tilde{p} \,\|\, p)$ is equivalent to maximizing the probability that the field $p$ assigns to the training data, given by

$$\prod_{1 \leq i \leq N} p(\omega^{(i)}) = \prod_{\omega \in \Omega} p(\omega)^{c(\omega)} \,\propto\, e^{-ND(\tilde{p} \,\|\, p)}. \tag{2.8}$$

With sufficiently many parameters it is a simple matter to construct a field for which $D(\tilde{p} \parallel p)$ is arbitrarily small. In fact, we can construct a field with $N+1$ features and small Kullback-Leibler divergence with respect to $\tilde{p}$ by taking

$$f_i(\omega) = \delta(\omega, \omega^{(i)}), \quad \lambda_i = \log c(\omega^{(i)}) \qquad (2.9)$$

for $1 \leq i \leq N$ and

$$f_{N+1}(\omega) = \prod_{1 \leq i \leq N} (1 - f_i(\omega)), \quad \lambda_{N+1} \ll -1. \qquad (2.10)$$

While such a model has small divergence with respect to the empirical distribution of the samples $\omega^{(i)}$, it does not generalize to other, previously unseen configurations. This is the classic problem of *over-training*. To avoid this problem we seek to incrementally construct a field that captures the salient properties of $\tilde{p}$ by incorporating an increasingly detailed collection of features. This motivates the random field induction paradigm that we now present.

*2.3 Inducing field interactions.* We begin by supposing that we have a set of *atomic* features

$$\mathcal{F}_{\text{atomic}} \subset \{g : \Omega \longrightarrow \{0, 1\}, \; \text{supp}(g) = v_g \in V\}$$

each of which is supported by a single vertex. We use atomic features to incrementally build up more complicated features. The following definition specifies how we shall allow a field to be incrementally constructed, or *induced*.

**Definition 2.1.** *Suppose that the field $q$ is given by $q = (\lambda \cdot f) \circ q_0$. The features $f_i$ are called the active features of $q$. A feature $g$ is a candidate for $q$ if either $g \in \mathcal{F}_{\text{atomic}}$, or if $g$ is of the form $g(\omega) = a(\omega) f_i(\omega)$ for an atomic feature $a$ and an active feature $f_i$ with $\text{supp}(g) \ominus \text{supp}(f_i) \in E$. The set of candidate features of $q$ is denoted $\mathcal{C}(q)$.*

In other words, candidate features are obtained by conjoining atomic features with existing features. The condition on supports ensures that each feature is supported by a path-connected subset of $G$. As an illustration, the figure below shows a situation in which the underlying graph $G$ is a grid, and a feature is supported by five vertices. The dashed lines indicate edges that would need to be present for the underlying field to be Markovian.

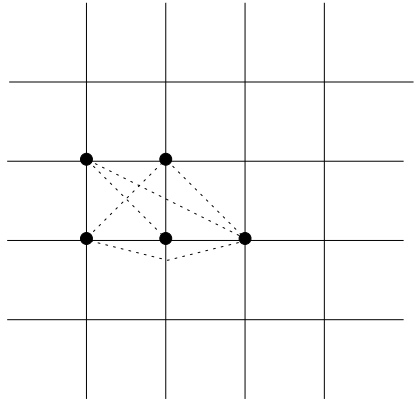

Figure 1

If $g \in \mathcal{C}(q)$ is a candidate feature of $q$, then we call the 1-parameter family of random fields $q_{\alpha g} = (\alpha g) \circ q$ the *induction of $q$ by $g$*. We also define

$$G_q(\alpha, g) = D(\tilde{p} \,\|\, q) - D(\tilde{p} \,\|\, q_{\alpha g}).  \qquad (2.11)$$

We think of $G_q(\alpha, g)$ as the improvement that feature $g$ brings to the model when it has weight $\alpha$. As we show in the following section, $G_q(\alpha, g)$ is $\cap$-convex in $\alpha$. We define $G_q(g)$ to be the greatest improvement that feature $g$ can give to the model while keeping all of the other features' parameters fixed:

$$G_q(g) = \sup_\alpha G_q(\alpha, g). \qquad (2.12)$$

We refer to $G_q(g)$ as the *gain* of the candidate $g$.

*2.4 Incremental construction of random fields.* We can now describe our algorithm for incrementally constructing fields.

**Field Induction Algorithm.**

    *Initial Data:*

        A reference distribution $\tilde{p}$ and an initial model $q_0$.

    *Output:*

        A field $q_\star$ with active features $f_0, \ldots, f_N$ such that $q_\star = \underset{q \in \mathcal{Q}(f, q_0)}{\arg\min}\, D(\tilde{p} \,\|\, q)$.

*Algorithm:*

(0) *Set $q^{(0)} = q_0$.*

(1) *For each candidate $g \in \mathcal{C}(q^{(n)})$ compute the gain $G_{q^{(n)}}(g)$.*

(2) *Let $f_n = \underset{g \in \mathcal{C}(q^{(n)})}{\arg\max}\, G_{q^{(n)}}(g)$ be the feature with the largest gain.*

(3) *Compute $q_\star = \underset{q \in \tilde{\mathcal{Q}}(f, q_0)}{\arg\min}\, D(\tilde{p} \,\|\, q)$, where $f = (f_0, f_1, \ldots, f_n)$.*

(4) *Set $q^{(n+1)} = q_\star$ and $n \leftarrow n + 1$, and go to step (1).*

This induction algorithm has two parts: *feature selection* and *parameter estimation*. Feature selection is carried out in steps (1) and (2), where the feature yielding the largest gain is incorporated into the model. Parameter estimation is carried out in step (3), where the parameters are adjusted to best represent the reference distribution. These two computations are discussed in more detail in the following two sections.

## 3. Feature Selection

The feature selection step of our induction algorithm is based upon an approximation. We assume that we can estimate the improvement due to adding a single feature, measured by the reduction in Kullback-Leibler divergence, by adjusting only the weight of the feature and keeping all of the other parameters of the field fixed. In general this is only an estimate, and it may well be that adding a feature will require significant adjustments to all of the parameters in the new model. From a computational perspective, approximating the improvement in this way can enable the simultaneous evaluation of thousands of candidate features, and makes the algorithm practical. In this section we present further detail on the feature selection step.

**Proposition 3.1.** *Let $G_q(\alpha, g)$, defined in (2.11), be the approximate improvement obtained by adding feature $g$ with parameter $\alpha$ to the field $q$. Then if $g$ is not constant, $G_q(\alpha, g)$ is strictly $\cap$-convex in $\alpha$ and attains its maximum at the unique point $\hat{\alpha}$ satisfying*

$$\tilde{p}[g] = q_{\hat{\alpha}g}[g]. \tag{3.1}$$

*Proof.* Using the definition (2.7) of the Kullback-Leibler divergence we can write

$$G_q(\alpha, g) = \sum_{\omega \in \Omega} \tilde{p}(\omega) \log \frac{Z_q^{-1}(\alpha g) e^{\alpha g(\omega)} q(\omega)}{q(\omega)}$$
$$= \sum_{\omega \in \Omega} \tilde{p}(\omega) \left( \alpha g(\omega) - \log q\left[e^{\alpha g}\right] \right) \qquad (3.2)$$
$$= \alpha \tilde{p}[g] - \log q\left[e^{\alpha g}\right].$$

Thus

$$\frac{\partial}{\partial \alpha} G_q(\alpha, g) = \tilde{p}[g] - \frac{q[g e^{\alpha g}]}{q[e^{\alpha g}]}$$
$$= \tilde{p}[g] - q_{\alpha g}[g].$$

Moreover,

$$\frac{\partial^2}{\partial \alpha^2} G_q(\alpha, g) = \frac{q[g e^{\alpha g}]^2}{q[e^{\alpha g}]^2} - \frac{q[g^2 e^{\alpha g}]}{q[e^{\alpha g}]}$$
$$= -q_{\alpha g}[(g - q_{\alpha g}[g])^2]$$

Hence, $\frac{\partial^2}{\partial \alpha^2} G_q(\alpha, g) \leq 0$, so that $G_q(\alpha, g)$ is ∩-convex in $\alpha$. If $g$ is not constant, then $\frac{\partial^2}{\partial \alpha^2} G_q(\alpha, g)$, which is minus the variance of $g$ with respect to $q_{\alpha g}$, is strictly negative, so that $G_q(\alpha, g)$ is strictly convex. □

When $g$ is binary-valued, its gain can be expressed in a particularly nice form. This is stated in the following proposition, whose proof is a simple calculation.

**Proposition 3.2.** *Suppose that the candidate $g$ is binary-valued. Then $G_q(\alpha, g)$ is maximized at*

$$\hat{\alpha} = \log \left( \frac{\tilde{p}[g](1 - q[g])}{q[g](1 - \tilde{p}[g])} \right) \qquad (3.3)$$

*and at this value,*

$$G_q(g) = G_q(\hat{\alpha}, g) = D(B_p \parallel B_q) \qquad (3.4)$$

*where $B_p$ and $B_q$ are Bernoulli random variables given by*

$$B_p(1) = \tilde{p}[g] \quad B_p(0) = 1 - \tilde{p}[g]$$
$$B_q(1) = q[g] \quad B_q(0) = 1 - q[g]. \qquad (3.5)$$

For features that are not binary-valued, but instead take values in the positive integers, the parameter $\hat{\alpha}$ that solves (3.1) and thus maximizes $G_q(\alpha, g)$ cannot, in general, be determined in closed form. This is the case for tied binary features, and it applies to the application we describe in Section 5. For these cases it is convenient to rewrite (3.1) slightly. Let $\beta = e^\alpha$ so that $\partial/\partial\alpha = \beta\partial/\partial\beta$. Let

$$g_k = \sum_\omega q(\omega)\,\delta(k, g(\omega)) \tag{3.6}$$

be the total probabilty assigned to the event that the feature $g$ takes the value $k$. Then (3.1) becomes

$$\beta\frac{\partial}{\partial\beta} G_q(\log\beta, g) = \tilde{p}[g] - \frac{\sum_{k=0}^N k\, g_k \beta^k}{\sum_{k=0}^N g_k \beta^k} = 0 \tag{3.7}$$

This equation lends itself well to numerical solution. The general shape of the curve $\beta \mapsto \beta\partial/\partial\beta\; G_q(\log\beta, g)$ is shown in the figure below.

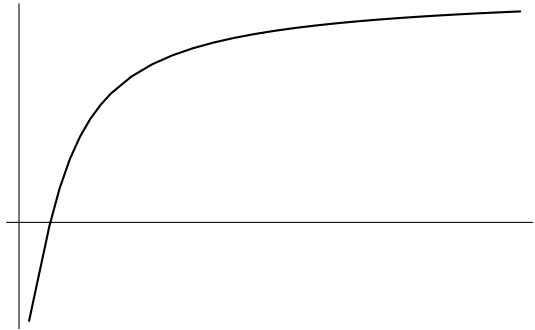

Figure 2

The limiting value of $\beta\partial G_q(\log\beta, g)/\partial\beta$ as $\beta \to \infty$ is $N - \tilde{p}[g]$. If $N - \tilde{p}[g] < 0$ then there is no solution to equation (3.7). Otherwise, the solution can be found using Newton's method, which in practice converges rapidly for such functions.

When the configuration space $\Omega$ is large, so that the coefficients $g_k$ cannot be calculated by summing over all configurations, Monte Carlo techniques may be used to estimate them. It is important to emphasize that the *same* set of random configurations can be used to estimate the coefficients $g_k$ for each candidate $g$ simultaneously. Rather than discuss the details of Monte Carlo techniques for this problem we refer to the extensive literature on this topic. We have obtained good results using the standard technique of Gibbs sampling [25] for the problem we describe in Section 5.

## 4. Parameter Estimation

In this section we present an algorithm for selecting the parameters associated with the features of a random field. The algorithm is closely related to the Generalized Iterative

Scaling algorithm of Darroch and Ratcliff [19]. Like the Darroch and Ratcliff procedure, the algorithm requires that the features $f_i$ are non-negative: $f_i(\omega) \geq 0$ for all $\omega \in \Omega$. Unlike the Darroch and Ratcliff procedure, however, our method does not require the features to be normalized to sum to a constant.

Throughout this section we hold the set of features $f = (f_0, f_1, \ldots, f_n)$, the initial model $q_0$ and the reference distribution $\tilde{p}$ fixed, and we simplify the notation accordingly. In particular, we write $\gamma \circ q$ instead of $(\gamma \cdot f) \circ q$ for $\gamma \in \mathbf{R}^n$. We assume that $q_0(\omega) = 0$ whenever $\tilde{p}(\omega) = 0$. This condition is commonly written $\tilde{p} \ll q_0$, and it is equivalent to $D(\tilde{p} \parallel q_0) < \infty$.

A description of the algorithm requires an additional piece of notation. Let

$$f_\#(\omega) = \sum_{i=0}^{n} f_i(\omega). \tag{4.1}$$

If the features are binary, then $f_\#(\omega)$ is the total number of features that are "on" for the configuration $\omega$.

**Improved Iterative Scaling.**

> *Initial Data:*
>
> > *A reference distribution $\tilde{p}$ and an initial model $q_0$, with $\tilde{p} \ll q_0$, and non-negative features $f_0, f_1, \ldots, f_n$.*
>
> *Output:*
>
> > *The distribution $q_\star = \underset{q \in \tilde{\mathcal{Q}}(f, q_0)}{\arg\min}\, D(\tilde{p} \parallel q)$*
>
> *Algorithm:*
>
> > *(0) Set $q^{(0)} = q_0$.*
> >
> > *(1) For each $i$ let $\gamma_i^{(k)} \in [-\infty, \infty)$ be the unique solution of*
> >
> > $$q^{(k)}[\, f_i\, e^{\gamma_i^{(k)}\, f_\#}\,] = \tilde{p}[\, f_i\,]. \tag{4.2}$$
> >
> > *(2) Set $q^{(k+1)} = \gamma^{(k)} \circ q^{(k)}$ and $k \leftarrow k + 1$.*
> >
> > *(3) If $q^{(k)}$ has converged, set $q_\star = q^{(k)}$ and terminate. Otherwise go to step (1).*

In other words, this algorithm constructs a distribution $q_\star = \lim_{n \to \infty} \gamma_n \circ q_0$ where $\gamma_n = \sum_{k=0}^{n} \gamma_i^{(k)}$ and $\gamma_i^{(k)}$ is determined as the solution to the equation

$$\sum_\omega q^{(k)}(\omega)\, f_i(\omega)\, e^{\gamma_i^{(k)}\, f_\#(\omega)} = \sum_\omega \tilde{p}(\omega)\, f_i(\omega). \tag{4.3}$$

When used in the $n$-th iteration of the field induction algorithm, where a candidate feature $g = f_n$ is added to the field $q = q_n$, we choose the initial distribution $q_0$ to be $q_0 = q_{\hat{\alpha}g}$, where $\hat{\alpha}$ is the parameter that maximizes the gain of $g$. In practice, this provides a good starting point from which to begin iterative scaling. In fact, we can view this distribution as the result of applying one iteration of an Iterative Proportional Fitting Procedure [8,15] to project $q_{\alpha g}$ onto the linear family of distributions with $g$-marginals constrained to $\tilde{p}[g]$.

Our main result in this section is

**Proposition 4.1.** *Suppose $q^{(k)}$ is the sequence in $\Delta$ determined by the Improved Iterative Scaling algorithm. Then $D(\tilde{p} \parallel q^{(k)})$ decreases monotonically to $D(\tilde{p} \parallel q_\star)$ and $q^{(k)}$ converges to $q_\star = \arg\min_{q \in \bar{\mathcal{Q}}} D(\tilde{p} \parallel q) = \arg\min_{p \in \mathcal{P}} D(p \parallel q_0)$.*

In the remainder of this section we present a self-contained proof of the convergence of the algorithm. The key idea of the proof is to express the incremental step of the algorithm in terms of an auxiliary function which bounds from below the likelihood objective function. This technique is the standard means of analyzing the EM algorithm [21], but it has not previously been applied to iterative scaling. Our analysis of iterative scaling is different and simpler than previous treatments. In particular, in contrast to Csiszár's proof of the Darroch-Ratcliff procedure [16], our proof does not rely upon the convergence of alternating I-projection [15].

We begin by proving the basic duality theorem which states that the maximum likelihood problem for a Gibbs distribution and the maximum entropy problem subject to linear constraints have the same solution. We then turn to the task of computing this solution. After introducing auxiliary functions in a general setting, we apply this method to prove convergence of the Improved Iterative Scaling algorithm. We finish the section by discussing Monte Carlo methods for estimating the equations when the size of the configuration space prevents the explicit calculation of feature expectations.

*4.1 Duality.* In this section we prove

**Proposition 4.2.** *Suppose that $\tilde{p} \ll q_0$. Then there exists a unique $q_\star \in \Delta$ satisfying*

 (1) $q_\star \in \mathcal{P} \cap \bar{\mathcal{Q}}$
 (2) $D(p \parallel q) = D(p \parallel q_\star) + D(q_\star \parallel q)$ *for any $p \in \mathcal{P}$ and $q \in \bar{\mathcal{Q}}$*
 (3) $q_\star = \arg\min_{q \in \bar{\mathcal{Q}}} D(\tilde{p} \parallel q)$
 (4) $q_\star = \arg\min_{p \in \mathcal{P}} D(p \parallel q_0)$.

*Moreover, any of these four properties determines $q_\star$ uniquely.*

This result is well known, although perhaps not quite in this packaging. In the language of constrained optimization, it expresses the fact that the maximum likelihood problem for Gibbs distributions is the convex dual to the maximum entropy problem for linear constraints. We include a proof here to make this paper self-contained and also to carefully address the technical issues arising from the fact that $\mathcal{Q}$ is not closed. The proposition would not be true if we replaced $\bar{\mathcal{Q}}$ with $\mathcal{Q}$. In fact, $\mathcal{P} \cap \mathcal{Q}$ might be empty. Our proof is elementary and does not rely on the Kuhn-Tucker theorem or other machinery of constrained optimization.

Our proof of the proposition will use a few lemmas. The first two lemmas we state without proof.

**Lemma 4.3.**
  (1) $D(p \,\|\, q)$ is a non-negative, extended real-valued function on $\Delta \times \Delta$.
  (2) $D(p \,\|\, q) = 0$ if and only if $p = q$.
  (3) $D(p \,\|\, q)$ is strictly convex in $p$ and $q$ separately.
  (4) $D(p \,\|\, q)$ is $C^1$ in $q$.

**Lemma 4.4.**
  (1) The map $(\gamma, p) \mapsto \gamma \circ p$ is smooth in $(\gamma, p) \in \mathbf{R}^n \times \Delta$.
  (2) The derivative of $D(p \,\|\, \lambda \circ q)$ with respect to $\lambda$ is

$$\frac{d}{dt}\Big|_{t=0} D(p \,\|\, (t\lambda) \circ q) = \lambda \cdot (p[f] - q[f]).$$

**Lemma 4.5.** If $\tilde{p} \ll q_0$ then $\mathcal{P} \cap \bar{\mathcal{Q}}$ is nonempty.

*Proof.* Define $q_\star$ by property (3) of Proposition 4.2; that is, $q_\star = \arg\min_{q \in \bar{\mathcal{Q}}} D(\tilde{p}, q)$. To see that this makes sense, note that since $\tilde{p} \ll q_0$, $D(\tilde{p}, q)$ is not identically $\infty$ on $\bar{\mathcal{Q}}$. Also, $D(p \,\|\, q)$ is continuous and strictly convex as a function of $q$. Thus, since $\bar{\mathcal{Q}}$ is closed, $D(\tilde{p}, q)$ attains its minimum at a unique point $q_\star \in \bar{\mathcal{Q}}$. We will show that $q_\star$ is also in $\mathcal{P}$. Since $\bar{\mathcal{Q}}$ is closed under the action of $\mathbf{R}^n$, $\lambda \circ q_\star$ is in $\bar{\mathcal{Q}}$ for any $\lambda$. Thus by the definition of $q_\star$, $\lambda = 0$ is a minimum of the function $\lambda \to D(\tilde{p}, \lambda \circ q_\star)$. Taking derivatives with respect to $\lambda$ and using Lemma 4.4 we conclude $q_\star[f] = \tilde{p}[f]$. Thus $q_\star \in \mathcal{P}$. □

**Lemma 4.6.** If $q_\star \in \mathcal{P} \cap \bar{\mathcal{Q}}$ then for any $p \in \mathcal{P}$ and $q \in \bar{\mathcal{Q}}$

$$D(p \,\|\, q) = D(p \,\|\, q_\star) + D(q_\star \,\|\, q).$$

This is called the *Pythagorean property* since it resembles the Pythagorean theorem if we imagine that $D(p \,\|\, q)$ is the square of Euclidean distance and $(p, q_\star, q)$ are the vertices of a right triangle.

*Proof.* A straightforward calculation shows that

$$D(p_1 \| q_1) - D(p_1 \| q_2) - D(p_2 \| q_1) + D(p_2 \| q_2) = \lambda \cdot (p_1[f] - p_2[f])$$

for any $p_1, p_2, q_1, q_2 \in \Delta$ with $q_2 = \lambda \circ q_1$. It follows from this identity and the continuity of $D$ that

$$D(p_1 \| q_1) - D(p_1 \| q_2) - D(p_2 \| q_1) + D(p_2 \| q_2) = 0$$

if $p_1, p_2 \in \mathcal{P}$ and $q_1, q_2 \in \bar{\mathcal{Q}}$. The lemma follows by taking $p_1 = q_1 = q_\star$. $\square$

*Proof of Proposition 4.2.* Choose $q_\star$ to be any point in $\mathcal{P} \cap \bar{\mathcal{Q}}$. Such a $q_\star$ exists by Lemma 4.5. It satisfies property (1) by definition, and it satisfies property (2) by Lemma 4.6. As a consequence of property (2), it also satisfies properties (3) and (4). To check property (3), for instance, note that if $q$ is any point in $\bar{\mathcal{Q}}$, then $D(\tilde{p} \| q) = D(\tilde{p} \| q_\star) + D(q_\star \| q) \geq D(q_\star \| q)$.

It remains to prove that each of the four properties (1)–(4) determines $q_\star$ uniquely. In other words, we need to show that if $m$ is any point in $\Delta$ satisfying any of the four properties (1)–(4), then $m = q_\star$. Suppose that $m$ satisfies property (1). Then by property (2) for $q_\star$ with $p = q = m$, $D(m, m) = D(m, q_\star) + D(q_\star, m)$. Since $D(m, m) = 0$, it follows that $D(m, q_\star) = 0$ so $m = q_\star$. If $m$ satisfies property (2), then the same argument with $q_\star$ and $m$ reversed again proves that $m = q_\star$. Suppose that $m$ satisfies property (3). Then

$$D(\tilde{p} \| q_\star) \geq D(\tilde{p} \| m) = D(\tilde{p} \| q_\star) + D(q_\star \| m)$$

where the second equality follows from property (2) for $q_\star$. Thus $D(q_\star, m) \leq 0$ so $m = q_\star$. If $m$ satisfies property (4), then a similar proof shows that once again $m = q_\star$. $\square$

*4.2 Auxiliary functions.* In the previous section we proved the existence of a unique probability distribution $q_\star$ that is both a maximum likelihood Gibbs distributions and a maximum entropy constrained distribution. We now turn to the task of computing $q_\star$.

Fix $\tilde{p}$ and let $L : \Delta \to \mathbf{R}$ be the log-likelihood objective function

$$L(q) = -D(\tilde{p} \| q).$$

**Definition 4.7.** *A function $A : \mathbf{R}^n \times \Delta \to R$ is an auxiliary function for $L$ if*

*(1) For all $q \in \Delta$ and $\gamma \in \mathbf{R}^n$*

$$L(\gamma \circ q) \geq L(q) + A(\gamma, q)$$

(2) $A(\gamma, q)$ is continuous in $q \in \Delta$ and $C^1$ in $\gamma \in \mathbf{R}^n$ with

$$A(0, q) = 0 \quad \text{and} \quad \frac{d}{dt}|_{t=0} A(t\gamma, q) = \frac{d}{dt}|_{t=0} L((t\gamma) \circ q).$$

We can use an auxiliary function $A$ to construct an iterative algorithm for maximizing $L$. We start with $q^{(k)} = q_0$ and recursively define $q^{(k+1)}$ by

$$q^{(k+1)} = \gamma^{(k)} \circ q^{(k)} \quad \text{with} \quad \gamma^{(k)} = \arg\max_\gamma A(\gamma, q^{(k)}).$$

It is clear from property (1) of the definition that each step of this procedure increases $L$. The following proposition implies that in fact the sequence $q^{(k)}$ will reach the maximum of $L$.

**Proposition 4.8.** *Suppose $q^{(k)}$ is any sequence in $\Delta$ with*

$$q^{(0)} = q_0 \quad \text{and} \quad q^{(k+1)} = \gamma^{(k)} \circ q^{(k)}$$

*where $\gamma^{(k)} \in \mathbf{R}^n$ satisfies*

$$A(\gamma^{(k)}, q^{(k)}) = \sup_\gamma A(\gamma, q^{(k)}). \tag{4.4}$$

*Then $L(q^{(k)})$ increases monotonically to $\max_{q \in \bar{\mathcal{Q}}} L(q)$ and $q^{(k)}$ converges to $q_\star = \arg\max_{q \in \bar{\mathcal{Q}}} L(q)$.*

Equation (4.4) assumes that the supremum $\sup_\gamma A(\gamma, q^{(k)})$ is achieved at finite $\gamma$. In the next section, under slightly stronger assumptions, we present a extension of Proposition 4.8 that allows some components of $\gamma^{(k)}$ to take the value $-\infty$.

To use the proposition to construct a practical algorithm we must determine an auxiliary function $A(\gamma, q)$ for which $\gamma^{(k)}$ satisfying the required condition can be determined efficiently. In the Section 4.3 we present a choice of auxiliary function which yields the Improved Iterative Scaling updates.

To prove Proposition 4.8 we first prove three lemmas.

**Lemma 4.9.** *If $m$ is a cluster point of $q^{(k)}$, then $A(\gamma, m) \leq 0$ for all $\gamma \in \mathbf{R}^n$.*

*Proof.* Let $q^{(k_l)}$ be a sub-sequence converging to $m$. Then for any $\gamma$

$$A(\gamma, q^{(k_l)}) \leq A(\gamma^{(k_l)}, q^{(k_l)}) \leq L(q^{(k_l+1)}) - L(q^{(k_l)}) \leq L(q^{(k_l+1)}) - L(q^{(k_l)}).$$

The first inequality follows from property (4.4) of $\gamma^{(n_k)}$. The second and third inequalities are a consequence of the monotonicity of $L(q^{(k)})$. The lemma follows by taking limits and using the fact that $L$ and $A$ are continuous. □

**Lemma 4.10.** *If $m$ is a cluster point of $q^{(k)}$, then $\frac{d}{dt}|_{t=0} L((t\gamma) \circ m) = 0$ for any $\gamma \in \mathbf{R}^n$.*

*Proof.* By the previous lemma, $A(\gamma, m) \leq 0$ for all $\gamma$. Since $A(0, m) = 0$, this means that $\gamma = 0$ is a maximum of $A(\gamma, m)$ so that

$$0 = \frac{d}{dt}|_{t=0} A(t\gamma, m) = \frac{d}{dt}|_{t=0} L((t\gamma) \circ m).$$

□

**Lemma 4.11.** *Suppose $\{q^{(k)}\}$ is any sequence with only one cluster point $q_*$. Then $q^{(k)}$ converges to $q_*$.*

*Proof.* Suppose not. Then there exists an open set $B$ containing $q_*$ and a subsequence $q^{(n_k)} \notin B$. Since $\Delta$ is compact, $q^{(n_k)}$ has a cluster point $q'_* \notin B$. This contradicts the assumption that $\{q^{(k)}\}$ has a unique cluster point. □

*Proof of Proposition 4.8.* Suppose that $m$ is a cluster point of $q^{(k)}$. It follows from Lemma 4.10 that $\frac{d}{dt}|_{t=0} L((t\gamma) \circ q) = 0$, and so $m \in \mathcal{P} \cap \bar{\mathcal{Q}}$ by Proposition 4.4. But $q_\star$ is the only point in $\mathcal{P} \cap \bar{\mathcal{Q}}$ by Proposition 4.2. It follows from Lemma 4.11 that $q^{(k)}$ converges to $q_\star$. □

*4.3 Dealing with $\infty$.* In order to prove the convergence of the Improved Iterative Scaling algorithm, we need an extension of Proposition 4.8 that allows the components of $\gamma$ to equal $-\infty$. For this extension, we assume that all the components of the feature function $f$ are non-negative:

$$f_i(\omega) \geq 0 \quad \text{for all } i \text{ and all } \omega. \tag{4.5}$$

Let $R \cup -\infty$ denote the partially extended real numbers with the usual topology. The operations of addition and exponentiation extend continuously to $R \cup -\infty$. Let $\mathcal{S}$ be the open subset of $(R \cup -\infty)^n \times \Delta$ defined by

$$\mathcal{S} = \{ (\gamma, q) \in (R \cup -\infty)^n \times \Delta : q(\omega) e^{\gamma \cdot f(\omega)} > 0 \text{ for some } \omega \}$$

Observe that $R^n \times \Delta$ is a dense subset of $\mathcal{S}$. The map $(\gamma, q) \mapsto \gamma \circ p$, which up to this point we defined only for finite $\gamma$, extends uniquely to a continuous map from all of $\mathcal{S}$ to $\Delta$. (The condition on $(\gamma, q) \in \mathcal{S}$ ensures that the normalization in the definition of $\gamma \circ p$ is well defined, even if $\gamma$ is not finite.)

**Definition 4.12.** *We call a function $A : \mathcal{S} \to R \cup -\infty$ an extended auxiliary function for $L$ if when restricted to $R^n \times \Delta$ it is an ordinary auxiliary function in the sense of Definition 4.7, and if, in addition, it satisfies property (1) of Definition 4.7 for any $(q, \gamma) \in \mathcal{S}$, even if $\gamma$ is not finite.*

Note that if an ordinary auxiliary function extends to a continuous function on $\mathcal{S}$, then the extension is an extended auxiliary function.

We have the following extension of Proposition 4.8:

**Proposition 4.8$'$.** *Suppose the feature function $f$ satisfies the non-negativity condition (4.5) and suppose $A$ is an extended auxiliary function for $L$. Then the conclusion of Proposition 4.8 continues to hold if the condition on $\gamma^{(k)}$ is replaced by:*

$$(\gamma^{(k)}, q^{(k)}) \in \mathcal{S} \quad \text{and} \quad A(\gamma^{(k)}, q^{(k)}) \geq A(\gamma, q^{(k)}) \quad \text{for any} \quad (\gamma, q^{(k)}) \in \mathcal{S}.$$

*Proof.* Lemma 4.9 is valid under the altered condition on $\gamma^{(k)}$ since $A(\gamma, q)$ satisfies property (1) of Definition 4.7 for all $(\gamma, q) \in \mathcal{S}$. As a consequence, Lemma 4.10 also is valid, and the proof of Proposition 4.8 goes through without change. □

*4.4 Improved Iterative Scaling.* We now prove the monotonicity and convergence of the Improved Iterative Scaling algorithm by applying Proposition 4.8 to a particular choice of auxiliary function. We continue to assume, as in the previous section, that each component of the feature function $f$ is non-negative.

For $q \in \Delta$ and $\gamma \in \mathbf{R}^n$, define

$$A(\gamma, q) = 1 + \gamma \cdot \tilde{p}[f] - \sum_\omega q(\omega) \sum_i f(i \,|\, \omega)\, e^{\gamma_i f_\#(\omega)}$$

where $f(i \,|\, \omega) = \frac{f_i(\omega)}{f_\#(\omega)}$. It is easy to check that $A$ extends to a continuous function on $(R \cup -\infty)^n \times \Delta$.

**Lemma 4.13.** *$A(\gamma, q)$ is an extended auxiliary function for $L(q)$.*

The key ingredient in the proof of the lemma is the ∩-convexity of the logarithm and the ∪-convexity of the exponential, as expressed in the inequalities

$$e^{\sum_i t_i \alpha_i} \leq \sum_i t_i\, e^{\alpha_i} \quad \text{if } t_i \geq 0 \text{ with } \sum_i t_i = 1 \tag{4.6}$$

$$\log x \leq x - 1 \quad \text{for all } x > 0. \tag{4.7}$$

*Proof of Lemma 4.13.* Because $A$ extends to a continuous function on $(R \cup -\infty)^n \times \Delta$, it suffices to prove that it satisfies properties (1) and (2) of Definition 4.7. To prove property (1) note that

$$L(\gamma \circ q) - L(q) = \gamma \cdot \tilde{p}[f] - \log \sum_\omega q(\omega) e^{\gamma \cdot f(\omega)} \tag{4.8}$$

$$\geq \gamma \cdot \tilde{p}[f] + 1 - \sum_\omega q(\omega) e^{\gamma \cdot f(\omega)} \tag{4.9}$$

$$\geq \gamma \cdot \tilde{p}[f] + 1 - \sum_\omega q(\omega) \sum_i f(i \mid \omega) e^{\gamma_i f_\#(\omega)} \tag{4.10}$$

$$= A(\gamma, q).$$

Equality (4.8) is a simple calculation. Inequality (4.9) follows from inequality (4.7). Inequality (4.10) follows from the definition of $f_\#$ and Jensen's inequality (4.6). Property (2) of Definition 4.7 is straightforward to verify. □

Proposition 4.1 follows immediately from the above lemma and the extended Proposition 4.8. Indeed, it is easy to check that $\gamma^{(k)}$ defined in Proposition 4.1 achieves the maximum of $A(\gamma, q^{(k)})$, so that it satisfies the condition of Proposition 4.8'.

*4.5 Monte Carlo methods.* The Improved Iterative Scaling algorithm described above is well-suited to numerical techniques since all of the features take non-negative values. In each iteration of this algorithm it is necessary to solve a polynomial equation for each feature $f_i$. That is, we can express equation (4.2) in the form

$$\sum_{m=0}^{M} a_{m,i}^{(k)} \beta_i^m = 0$$

where $M$ is the largest value of $f_\#(\omega) = \sum_i f_i(\omega)$ and

$$a_{m,i}^{(k)} = \begin{cases} \sum_\omega q^{(k)}(\omega) f_i(\omega) \delta(m, f_\#(\omega)) & n > 0 \\ -\tilde{p}[f_i] & m = 0 \end{cases} \tag{4.11}$$

where $q^{(k)}$ is the field for the $k$-th iteration and $\beta_i = e^{\gamma_i^{(k)}}$. This equation has no solution precisely when $a_{m,i}^{(k)} = 0$ for $m > 0$. Otherwise, it can be efficiently solved using Newton's method since all of the coefficients $a_{m,i}^{(k)}$, $m > 0$, are non-negative. When Monte Carlo methods are to be used because the configuration space $\Omega$ is large, the coefficients $a_{m,i}^{(k)}$ can be simultaneously estimated for all $i$ and $m$ by generating a single set of samples from the distribution $q^{(k)}$.

# 5. Application: Word Morphology

Word clustering algorithms are useful for many natural language processing tasks. One such algorithm [10], called mutual information clustering, is based upon the construction of simple bigram language models using the maximum likelihood criterion. The algorithm gives a hierarchical binary classification of words that has been used for a variety of purposes, including the construction of decision tree language and parsing models [28], and sense disambiguation for machine translation [11].

A fundamental shortcoming of the mutual information word clustering algorithm given in [10] is that it takes as fundamental the word spellings themselves. This increases the severity of the problem of small counts that is present in virtually every statistical learning algorithm. For example, the word "Hamiltonianism" appears only once in the 365,893,263-word corpus used to collect bigrams for the clustering experiments described in [10]. Clearly this is insufficient evidence on which to base a statistical clustering decision. The basic motivation behind the feature-based approach is that by querying features of spellings, a clustering algorithm could notice that such a word begins with a capital letter, ends in "ism" or contains "ian," and profit from how these features are used for other words in similar contexts.

In this section we describe how we applied the random field induction algorithm to discover morphological features of words, and we present sample results. This technique was used in [27] to improve mutual information clustering. In Section 5.1 we formlate the problem in terms of the notation and results of Sections 2, 3, and 4. In Section 5.2 we describe how the field induction algorithm is actually carried out in this application. In Section 5.3 we explain the results of the induction algorithm by presenting a series of examples.

*5.1 Problem formulation.* To discover features of spellings we take as configuration space $\Omega = \mathcal{A}^*$ where $\mathcal{A}$ is the ASCII alphabet. We construct a probability distribution $p(\omega)$ on $\Omega$ by first predicting the length $|\omega|$, and then predicting the actual spelling; thus, $p(\omega) = p_l(|\omega|)p_s(\omega | |\omega|)$ where $p_l$ is the length distribution and $p_s$ is the spelling distribution. We take the length distribution as given. We model the spelling distribution $p_s(\cdot | l)$ over strings of length $l$ as a random field. Let $\Omega_l$ be the configuration space of all ASCII strings of length $l$. Then $|\Omega_l| = O(10^{2l})$ since each $\omega_i$ is an ASCII character.

To reduce the number of parameters, we tie features so that a feature has the same weight independent of where it appears in the string. Because of this it is natural to view the graph underlying $\Omega_l$ as a regular $l$-gon. The group of automorphisms of this graph is the set of all rotations, and the resulting field is homogeneous as defined in Section 2.

Not only is each field $p_l$ homogeneous, but in addition, we tie features across fields for different values of $l$. Thus, the weight $\lambda_f$ of a feature is independent of $l$. To introduce a dependence on the length, as well as on whether or not a feature applies at the beginning or end of a string, we adopt the following artificial construction. We take as the graph of $\Omega_l$ an $(l+1)$-gon rather than an $l$-gon, and label a distinguished vertex by the length, keeping this label held fixed. The graph for a 7-letter word is depicted in Figure 3. The dashed lines indicate edges that would need to be present for the field to be Markovian if each feature is supported by no more than three vertices.

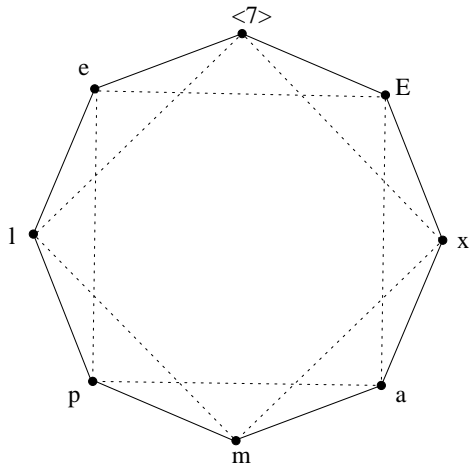

Figure 3

To complete the description of the fields that are induced, we need to specify the set of atomic features. The atomic features that we allow fall into three types. The first type is the class of features of the form

$$f_{v,c}(\omega) = \begin{cases} 1 & \text{if } \omega_v = c \\ 0 & \text{otherwise.} \end{cases}$$

where $c$ is any ASCII character. The second type of atomic features involve the special vertex <$l$> that carries the length of the string. These are the features

$$f_{v,l}(\omega) = \begin{cases} 1 & \text{if } \omega_v = \text{<}l\text{>} \\ 0 & \text{otherwise} \end{cases}$$

$$f_{v,\text{<>}}(\omega) = \begin{cases} 1 & \text{if } \omega_v = \text{<}l\text{> for some } l \\ 0 & \text{otherwise} \end{cases}$$

The atomic feature $f_{v,\text{<>}}$ introduces a dependence on whether a string of characters lies at the beginning or end of the string, and the atomic features $f_{v,l}$ introduce a dependence on the length of the string. To tie together the length dependence for long strings, we also introduce an atomic feature $f_{v,7+}$ for strings of length 7 or greater.

The final type of atomic feature asks whether a character lies in one of three sets, [a-z], [A-Z], [0-9], [@-&], denoting arbitrary lowercase letters, uppercase letters, digits, and punctuation. For example, the atomic feature

$$f_{v,\texttt{[a-z]}}(\omega) = \begin{cases} 1 & \text{if } \omega_v \in \texttt{[a-z]} \\ 0 & \text{otherwise} \end{cases}$$

tests whether or not a character is lowercase.

To illustrate the notation that we use, let us suppose that the the following features are active for a field: "ends in ism," "a string of at least 7 characters beginning with a capital letter" and "contains ian." Then the probability of the word "Hamiltonianism" would be given as

$$P_l(14)\,p_s(\texttt{Hamiltonianism}\,|\,|\omega| = 14) = P_l(14)\,\frac{1}{Z_{14}}\,e^{\lambda_{\texttt{7+<[A-Z]}} + \lambda_{\texttt{ian}} + \lambda_{\texttt{ism>}}}.$$

Here the $\lambda$'s are the parameters of the appropriate features, and we use the characters < and > to denote the beginning and ending of a string (more common regular expression notation would be ^ and $). The notation 7+<[A-Z] thus means "a string of at least 7 characters that begins with a capital letter," corresponding to the feature

$$f_{v,7+}\,f_{v,\texttt{[A-Z]}}\,.$$

Similarly, ism> means "ends in -ism" and corresponds to the feature

$$f_{v,\texttt{i}}\,f_{v,\texttt{s}}\,f_{v,\texttt{m}}\,f_{v,\texttt{<>}}$$

and ian means "contains ian," corresponding to the feature

$$f_{v,\texttt{i}}\,f_{v,\texttt{a}}\,f_{v,\texttt{n}}\,.$$

*5.2 Description of the algorithm.* We begin the random field induction algorithm with a model that assigns uniform probability to all word strings. We then incrementally add features to a random field model in order to minimize the Kullback-Leibler divergence between the field and the unigram distribution of the vocabulary obtained from a training corpus. The length distribution is taken according to the lengths of words in the empirical distribution of the training data. The improvement to the model made by a candidate feature is evaluated by the reduction in relative entropy, with respect to the unigram distribution, that adding the new feature yields, keeping the other parameters of the model fixed. Our learning algorithm incrementally constructs a random field to describe those features of spellings that are most informative.

At each stage in the induction algorithm, a set of candidate features is constructed. Because the fields are homogeneous, the set of candidate features can be viewed as follows. Each active feature can be expressed in the form

$$f_s(\omega) = \begin{cases} 1 & \text{substring } s \text{ appears in } \omega \\ 0 & \text{otherwise} \end{cases}$$

where $s$ is a string in the extended alphabet $\mathcal{A}$ of ASCII characters together with the macros [a-z], [A-Z], [0-9], [@-&], and the length labels <1> and <>. If $\{f_s\}_{s \in \mathcal{S}}$ is the set of active features, (including $s = \epsilon$) using this representation, then the set of candidate features is precisely the set

$$\{f_{a \cdot s}, f_{s \cdot a}\}_{a \in \mathcal{A}, s \in \mathcal{S}}$$

where $a \cdot s$ denotes concatenation of strings. As required by Definition 2, each such candidate increases the support of an active feature by a single adjacent vertex.

Since the model assigns probability to arbitrary word strings, the partition function $Z_l$ can be computed exactly for only the smallest string lengths $l$. We therefore compute feature expectations using a random sampling algorithm. Specifically, we use the *Gibbs sampler* [25] to generate 10,000 spellings of random lengths. When computing the gain $G_q(g)$ of a candidate feature, we use these spellings to estimate the probability $g_k$ that the candidate feature $g$ occurs $k$ times in a spelling (see equation (3.7)–for example, the feature $f_{v,[a-z]}$ occurs 2 times in the string The), and then solve for the corresponding $\beta$ using Newton's method for *each* candidate feature. It should be emphasized that only a single set of random spellings needs to be generated; the same set can be used to estimate $g_k$ for each candidate $g$. After adding the best candidate to the field, all of the feature weights are readjusted using the Improved Iterative Scaling algorithm. To carry out this algorithm, random spellings are again generated, this time incorporating the new feature, yielding Monte Carlo estimates of the coefficients $a_{m,i}^{(k)}$. Recall that $a_{m,i}^{(k)}$ is the expected number of times that feature $i$ appears (under the substring representation for homogeneous features) in a string for which there is a total of $m$ active features (see equation (4.11)). Given estimates for these coefficients, Newton's method is again used to solve equation (4.11), to complete a single iteration of the iterative scaling algorithm. After convergence of the Kullback-Leibler divergence, the inductive step is complete, and a new set of candidate features is considered.

*5.3 Sample results.* We began with a uniform field, that is, a field with no features at all. For this field, all ASCII strings of a given length are equally likely, and the lengths are drawn from a fixed distribution. Here is a sample of strings drawn from this distribution:

```
~, mo, _!ZP*@, m/TLL, ks;cm_3, *LQdR, D, aW{, 5&TL|4, tc, ?!@,
sNeiO+, wHo8zBr", pQlV, m, H!&, h9, #Os, :, Ky}FM?, LW, ",8},
89Lj, -P, A, !, H, ', Y^:Du:, 1xCl, 1!'J#F*u., w=idHnM), ~, 2,
2leW2, I,bw~tk1, 3", ], ], b, +JEmj6, +E*, \qjqe"-7f, |al2, T,
~(sOc1+2ADe, &, \p9oH, i;, $6, q}O+[, xEv, #U, O)[83COF,
=|B|7%cR, Mqq, ?!mv, n=7G, $i9GAJ\, D, 5, ,=, +u6@I9:, +, =D,
2E#vz@3-, ~nu;.+s, 3xJ, GDWeqL, R,3R, !7v, FX,@y, 4p_cY2hU, ~
```

It comes as no surprise that the first feature the induction algorithm chooses is [a-z]; it simply observes that characters should be lowercase. The maximum likelihood (maximum entropy) weight for this feature is $\beta = e^\lambda \approx 6.99$. This means that a string with a lowercase letter in some position is about 7 times more likely than the same string without a lowercase letter in that position.

When we now draw strings from the new distribution (using annealing to concentrate the distribution on the more probable strings), we obtain spellings that are primarily made up of lowercase letters, but that certainly do not resemble English words:

```
m, r, xevo, ijjiir, b, to, jz, gsr, wq, vf, x, ga, msmGh,
pcp, d, oziVlal, hzagh, yzop, io, advzmxnv, ijv_bolft, x,
emx, kayerf, mlj, rawzyb, jp, ag, ctdnnnbg, wgdw, t, kguv,
cy, spxcq, uzflbbf, dxtkkn, cxwx, jpd, ztzh, lv, zhpkvnu,
l^, r, qee, nynrx, atze4n, ik, se, w, lrh, hp+, yrqyka'h,
zcngotcnx, igcump, zjcjs, lqpWiqu, cefmfhc, o, lb, fdcY, tzby,
yopxmvk, by, fz,, t, govyccm, ijyiduwfzo, 6xr, duh, ejv, pk,
pjw, l, fl, w
```

In the following table we show the first 10 features that the algorithm induced, together with their associated parameters. Several things are worth noticing. The second feature chosen was [a-z][a-z], which denotes adjacent lowercase characters. The third feature added was the letter e, which is the most common letter. The weight for this feature is $\beta = e^\lambda = 3.47$. The next feature introduces the first dependence on the length of the string: `[a-z]>1` denotes the feature "a one character word ending with a lowercase letter." Notice that this feature has a small weight of 0.04, corresponding to our intuition that such words are uncommon. Similarly, the features z, q, j, and x are uncommon, and thus receive small weights. The appearance of the feature * is explained by the fact that the vocabulary for our corpus is restricted to the most frequent 100,000 spellings, and all other words receive the "unknown word" spelling ***, which is rather frequent. (The "end-of-sentence" marker, which makes its appearance later, is given the spelling |.)

| feature | [a-z] | [a-z][a-z] | e | [a-z]>1 | t | * | z | q | j | x |
|---|---|---|---|---|---|---|---|---|---|---|
| $\beta$ | 6.64 | 6.07 | 3.47 | 0.04 | 2.75 | 17.25 | 0.02 | 0.03 | 0.02 | 0.06 |

Shown below is a collection of spellings obtained by Gibbs sampling from the resulting collection fields.

```
frk, et, egeit, edet, eutdmeeet, ppge, A, dtgd, falawe, etci,
eese, ye, epemtbn, tegoeed, ee, *mp, temou, enrteunt, ore,
erveelew, heyu, rht, *, lkaeu, lutoee, tee, mmo, eobwtit,
weethtw, 7, ee, teet, gre, /, *, eeeteetue, hgtte, om, he, *,
stmenu, ec, ter, eedgtue, iu, ec, reett, *, ivtcmeee, vt, eets,
tidpt, lttv, *, etttvti, ecte, X, see, *, pi, rlet, tt, *, eot,
leef, ke, *, *, tet, iwteeiwbeie, yeee, et, etf, *, ov
```

After inducing 100 features, the model finally begins to be concentrated on spellings that resemble actual spellings to some extent, particularly for short words. At this point the algorithm has discovered, for example, that `the` is a very common 3-letter word, that many words end in `ed`, and that long words often end in `ion`. A sample of 10 of the first 100 features induced, with their appropriate weights is shown in the table below.

| feature | . | ,>1 | 3<the | tion | 4<th | y> | ed> | ion>7+ | ent | 7+<c |
|---|---|---|---|---|---|---|---|---|---|---|
| $\beta$ | 22.36 | 31.30 | 11.05 | 5.89 | 4.78 | 5.35 | 4.20 | 4.83 | 5.17 | 5.37 |

```
thed, and, thed, toftion, |, ieention, cention, |, ceetion,
ant, is, seieeet, cinention, and, ., tloned, uointe, feredten,
iined, sonention, inathed, other, the, id, and, ,, of, is, of,
of, ,, lcers, ,, ceeecion, ,, roferented, |, ioner, ,, |, the,
the, the, centention, ionent, asers, ,, ctention, |, of, thed,
of, uentie, of, and, ttentt, in, rerey, and, |, sotth, cheent,
is, and, of, thed, rontion, that, seoftr
```

A sample of the first 1000 features induced is shown in the table below, together with randomly generated spellings. Notice, for example, that the feature [0-9][0-9] appears with a surprisingly high weight of 9382.93. This is due to the fact that if a string contains one digit, then it's very likely to contain two digits. But since digits are relatively rare in general, the feature [0-9] must have a small weight of 0.038. Also, according to the model, a lowercase letter followed by an uppercase letter is rare.

| $feature$ | s>      | <re   | ght>  | 3<[A-Z] | ly>  | al>7+  | ing>      |
|-----------|---------|-------|-------|---------|------|--------|-----------|
| $\beta$   | 7.25    | 4.05  | 3.71  | 2.27    | 5.30 | 94.19  | 16.18     |
| $feature$ | [a-z][A-Z] | 't> | ed>7+ | er>7+  | ity  | ent>7+ | [0-9][0-9] |
| $\beta$   | 0.003   | 138.56| 12.58 | 8.11    | 4.34 | 6.12   | 9382.93   |
| $feature$ | qu      | ex    | ae    | ment    | ies  | <wh    | ate       |
| $\beta$   | 526.42  | 5.265 | 0.001 | 10.83   | 4.37 | 5.26   | 9.79      |

```
was, reaser, in, there, to, will, ,, was, by, homes, thing,
be, reloverated, ther, which, conists, at, fores, anditing, with,
Mr., proveral, the, ,, ***, on't, prolling, prothere, ,, mento,
at, yaou, 1, chestraing, for, have, to, intrally, of, qut, .,
best, compers, ***, cluseliment, uster, of, is, deveral, this,
thise, of, offect, inatever, thifer, constranded, stater, vill,
in, thase, in, youse, menttering, and, ., of, in, verate, of, to
```

Finally, we visit the state of the model after growing 1500 features to describe words. At this point the model is making more refined judgements regarding what is to be considered a word and what is not. The appearance of the features {}> and \[@-&]{, is explained by the fact that in preparing our corpus, certain characters were assigned special "macro" strings. For example, the punctuation characters $, _, %, and & are represented in our corpus as \${}, \_{}, \%{}, and \&{}. As the following sampled spellings demonstrate, the model has at this point recognized the existence of macros, but has not yet discerned their proper use.

| feature | 7+<inte | prov | <der | <wh | 19 | ons>7+ | ugh | ic> |
|---------|---------|------|------|-----|-----|--------|-----|-----|
| $\beta$ | 4.23 | 5.08 | 0.03 | 2.05 | 2.59 | 4.49 | 5.84 | 7.76 |
| feature | sys | ally | 7+<con | ide | nal | {}> | qui | \[@-&]{ |
| $\beta$ | 4.78 | 6.10 | 5.25 | 4.39 | 2.91 | 120.56 | 18.18 | 913.22 |
| feature | iz | IB | <inc | <im | iong | $ | ive>7+ | <un |
| $\beta$ | 10.73 | 10.85 | 4.91 | 5.01 | 0.001 | 16.49 | 2.83 | 9.08 |

```
the, you, to, by, conthing, the, ., not, have, devened, been,
of, |, F., ., in, have, -, ,, intering, ***, ation, said,
prouned, ***, suparthere, in, mentter, prement, intever, you,
., and, B., gover, producits, alase, not, conting, comment,
but, |, that, of, is, are, by, from, here{}, incements, contive,
., evined, agents, and, be, \.{}, thent, distements, all, --,
has, will, said, resting, had, this, was, intevent, IBM,
whree, acalinate, herned, are, ***, O., |, 1980, but, will,
***, is, ., to, becoment, ., with, recall, has, |, nother,
ments, was, the, to, of, stounicallity, with, camanfined,
in, this, intations, it, conanament, out, they, you
```

While clearly the model still has much to learn, it has at this point compiled a significant collection of morphological observations, and has traveled a long way toward its goal of statistically characterizing English spellings.

## 6. Extensions and Relations to Other Approaches

In this section we briefly discuss some relations between our incremental feature induction algorithm for random fields and other statistical learning paradigms. We also present some possible extensions and improvements of our method.

*6.1 Boltzmann machines.* There is an immediate resemblance between the parameter estimation problem for the random fields that we have considered and the learning problem for Boltzmann machines [1]. The classical Boltzmann machine is considered to be a random field on a graph $G = (E, V)$ with configuration space $\Omega = \{0, 1\}^V$ consisting of all labelings of the vertices by either a zero or a one. The machine is specified by a probability distribution on this configuration space of the form

$$p(\omega) = \frac{1}{Z} e^{\sum_{i,j} \lambda_{i,j} \omega_i \omega_j}$$

and the learning problem is to determine the set of weights $\lambda_{i,j}$ that best characterize a set of training samples. Typically only a subset of the vertices are labeled in the training set; the remaining vertices are considered to comprise the *hidden units*. Treated as a maximum likelihood problem, the training problem for Boltzmann machines becomes an instance of the general problem addressed by the EM algorithm [21], where iterative scaling is carried out in the M-step [12].

Most often the architecture of a Boltzmann machine is prescribed, and the learning problem is then solved by applying the EM algorithm (which typically involves random sampling and annealing). To cast Boltzmann machines into our framework, we can simply take binary-valued features of the form $f_{i,j}(\omega) = \omega_i \omega_j$. More generally, by allowing binary-valued features of the form

$$f_{\mathbf{v}}(\omega) = \omega_{v_1} \omega_{v_2} \cdots \omega_{v_n}$$

for $\mathbf{v} = (v_1, \ldots, v_n)$ a path in $G = (E, V)$, we construct models that are essentially "higher-order" Boltzmann machines [32]. With candidate features of this form our algorithm incrementally constructs a Boltzmann machine with no hidden units. If only a subset of the vertices are labelled in the training samples, then our Improved Iterative Scaling algorithm becomes an instance of the EM algorithm.

*6.2 Decision trees.* Our feature induction method also bears some resemblance to various methods for growing decision trees. Like decision trees, our method builds a top-down classification that refines features. However, decision trees correspond to constructing features that have disjoint support. For example, binary decision trees are grown by splitting a mode $n$ into two nodes by asking a binary question $q_n$ of the data at that node. Questions can be evaluated by the amount by which they reduce the entropy of the data at that node. This corresponds to our criterion of maximizing the reduction in entropy $G_q(g)$ over all candidate features $g$ for a field $q$. When the decision tree has been grown to completion, each leaf $l$ corresponds to a sequence of binary features

$$f_l, \; f_{l\uparrow}, \; f_{l\uparrow\uparrow}, \; \ldots, \; f_{\text{root}}$$

where $n\uparrow$ denotes the parent of node $n$, and with each feature $f_n$ being either the question $q_n$ or its negation $\neg q_n$. Thus, each leaf $l$ is characterized by the conjunction of these features, and different leaves correspond to conjunctions with disjoint support. In contrast, our feature induction algorithm generally results in features that have overlapping support.

By modifying our induction algorithm in the following way, we obtain a direct generalization of binary decision trees. Instead of considering the 1-parameter family of fields $q_{\lambda,g}$ to determine the best candidate $g = a \wedge f$, we consider the 2-parameter family of fields given by

$$q_{\lambda,\lambda',g} = \frac{1}{Z_{\lambda,\lambda',g}} e^{\lambda a \wedge f + \lambda'(\neg a) \wedge f}.$$

Since the features $a \wedge f$ and $(\neg a) \wedge f$ have disjoint support, the improvement obtained by adding both of them is given by $G_q(a \wedge f) + G_q((\neg a) \wedge f)$. This procedure generalizes decision trees since the resulting features in the field can be overlapping.

*6.3 Dynamic Markov coding.* Another technique that is similar in some aspects to random field induction is the *dynamic Markov coding* technique for text compression [14,5]. To incrementally build a finite state machine for generating strings in some output alphabet, dynamic Markov coding is based on the heuristic that the relative entropy of the finite-state machine might be lowered by giving a unique destination state to arcs that have high count. At each stage in the algorithm a state in the machine is split, or "cloned," into two states. The arc with the highest count coming into the original state is attached to one of the new states, and all of the remaining input arcs are attached to the other new state. As shown in [5], this technique is equivalent to incrementally building a finite-context model, adding a single output symbol $s$ to a valid prefix $\alpha$ to form a new valid prefix $s \cdot \alpha$. In this way it is similar to our field induction algorithm which at each stage generates a new feature of the form $a \wedge f$ for some active feature $f$. However, because of the "on-line" nature of dynamic Markov coding, the technique is unable to precisely calculate the reduction in entropy due to splitting a state, and must instead rely on more primitive heuristics. A closely related technique is given in [31]. In [33] a method for building hidden Markov models is presented which is in some sense the opposite approach, in that it starts with a maximally detailed finite-state model and proceeds by incrementally generalizing by merging states according to a greedy algorithm.

*6.4 Conditional exponential models.* Almost all of what we have presented here carries over to the more general setting of conditional exponential models, including the Improved Iterative Scaling algorithm presented here. For general conditional distributions $p(y \mid x)$ there may be no underlying random field, but with features defined as binary functions $f(x, y)$, the same general approach is applicable. The feature induction method for conditional exponential models is demonstrated for several problems in statistical machine translation in [6], where it is presented in terms of the principle of maximum entropy.

*6.5 Extensions.* The random field induction method presented in this paper is not definitive; there are many possible variations on the basic theme, which is to incrementally construct an increasingly detailed field to approximate the reference distribution $\tilde{p}$. Because the basic technique is based on a greedy algorithm, there are of course many ways for improving the search for a good set of features. The algorithm presented in Section 2 is in some respects the most simple possible within the general framework. But it also computationally intensive. A natural modification would be to add several of the top candidates at each stage. While this should increase the overall speed of the induction algorithm, it would also potentially result in more redundancy among the features, since the top candidates could be correlated. Another modification of the algorithm would be to add only the best candidate at each step, but then to carry out parameter estimation only after several new features had been added to the field. It would also be natural to establish a more Bayesian framework in which a prior distribution on features and parameters is incorporated. This could enable a principled approach for deciding when the feature induction is complete, by evaluating the posterior distribution of the field given the training samples.

As mentioned above, the method presented here does not explicitly learn any hidden structure, and thus does not generalize as much as would be desirable for many applications. One possibility would be to combine our method with a merging technique for combining features in order to generalize from a more detailed set of observations. While in principle our learning method can be carried out in the presence of incomplete data (in which case iterative scaling of the parameters can be viewed as an EM algorithm), we have not investigated searching methods for revealing hidden structure. This is a promising direction for future research.